%% file: main.tex
\documentclass[10pt,letterpaper]{article}

\usepackage[pagenumbers]{cvpr} % To force page numbers, e.g. for an arXiv version
\usepackage{listings}
\usepackage{float}
\usepackage{wrapfig}

\input{preamble}

\definecolor{cvprblue}{rgb}{0.21,0.49,0.74}
\usepackage[pagebackref,breaklinks]{hyperref}

\title{A Power Transform}

\author{Jonathan T. Barron \\
{\tt\small barron@google.com}
}

\begin{document}
\maketitle

Power transforms, such as the Box-Cox transform~\cite{BoxCox1964jrssB} and Tukey's ladder of powers~\cite{Tukey1977}, are a fundamental tool in mathematics and statistics. These transforms are primarily used for normalizing and standardizing datasets, effectively by raising values to a power. In this work I present a novel power transform, and I show that it serves as a unifying framework for wide family of loss functions, kernel functions, probability distributions, bump functions, and neural network activation functions.

\vspace{0.05in}

First, some history. Years ago I realized that many discrete robust loss functions in the literature were special cases of a single-parameter general robust loss function~\cite{barron2019general}. Later on, I realized that those loss functions could be framed as the result of a specific power transform applied to a quadratic loss function, and that this power transform was a useful tool in itself~\cite{barron2023zipnerf}. I have since realized that this power transform can be improved, and that it also unifies a variety of other individual concepts into useful single-parameter families. This paper is meant to formalize this unified framework, in the hopes that the community finds it useful. I have also identified and fixed numerical stability and speed issues with previous iterations of this power transform, and I present a fast and stable algorithm alongside a reference implementation.

\vspace{0.05in}

In Section~\ref{sec:power_transform} I will present an improved version of the previous power transform that can express a wider family of functions, and that has a symmetry that makes the transform ``self-inverting,'' which is a property I have found to be convenient.
In Section~\ref{sec:loss_function} I will show how this power transform generates a family of robust loss functions that is slightly wider than the family presented in my prior work~\cite{barron2019general}. In Section~\ref{sec:kernel_function} I will show how this transform can be used to generate a family of kernel functions that includes half-dozen previously-explored discrete kernel functions. In Sections~\ref{sec:prob}
and \ref{sec:bump} I will show how exponentiating the robust loss function yields both a family of probability distributions and a family of bump functions, both of which include many prior discrete distributions/functions as special cases. In Section~\ref{sec:neg_act} I will present a two-parameter power transform for positive and negative inputs which can be used to reproduce a number of commonly-used activation functions used by the deep learning community. In Section~\ref{sec:numerical} I will identify a problem with the numerical stability of this and prior power transforms, and propose a solution, and in Section~\ref{sec:implementation} I will present a fast, accurate, and stable implementation of this power transform, from which the aforementioned families of loss functions, kernel functions, probability distributions, bump functions, and activation functions can be trivially implemented.

\section{Power Transforms}
\label{sec:power_transform}

Here is the ``root'' form of the power transform I will be discussing. Everything else in this paper will build upon this.
\begin{equation}
\powerfun(x, \power) \triangleq \frac{2 \lft| \power \rgt|\,\,\,}{2 - \lft| \power \rgt| + \power} \cdot \lft( \lft( 1 + \frac{2 - \lft| \power \rgt| - \power}{2 \lft| \power \rgt|\,\,\,} x \rgt)^{\displaystyle \lft( 1 - \lft| \power \rgt| \rgt)^{-\operatorname{sgn} \lft( \power \rgt)}} - 1 \rgt)
\label{eq:f_root}
\end{equation}
This equation has some removable singularities and limits that can be filled in:
\begin{equation}
\powerfun(x, \power) = \lft\{ \begin{array}{ll}
-\log \lft( 1 - x \rgt) & \mathrm{if} \ \power = +\infty \\
\exp(x) - 1 & \mathrm{if} \ \power = 1 \\
\power \cdot \lft( \sqrt[1 - \power]{1 + \frac{1 - \power}{\power} x} - 1 \rgt) & \mathrm{if} \ 0 < \power < +\infty \land \power \ne 1 \\
x & \mathrm{if} \ \power = 0 \\
-\frac{\power}{\power + 1} \cdot \lft(  \lft( 1 - \frac{1}{\power} x \rgt)^{\power + 1} - 1 \rgt) & \mathrm{if} \ -\infty < \power < 0 \land \power \ne -1 \\
\log \lft( 1 + x \rgt) & \mathrm{if} \ \power = -1 \\ 
1 - \exp \lft( -x \rgt) & \mathrm{if} \ \power = -\infty
\end{array} \rgt.
\label{eq:f_unstable}
\end{equation}
There are other non-singular values for $\lambda$ that yield familiar forms of $f(x, \lambda)$:
\begin{align}
    f(x, -3) &= -\frac{3}{2} \cdot \lft( \frac{1}{\lft( 1 + \sfrac{x}{3} \rgt)^{2}} - 1 \rgt) \,, & \quad\quad f(x, 3) &= 3 \cdot \lft( \frac{1}{\sqrt{ 1 - \sfrac{2x}{3} }} - 1 \rgt)\,, \\
    \powerfun(x, -2) &= 2 \cdot \lft( 1 - \frac{1}{1 + \sfrac{x}{2}} \rgt) = \frac{2x}{x + 2}\,, & \quad\quad \powerfun(x, 2) &= 2 \cdot \lft( \frac{1}{1 - \sfrac{x}{2}} - 1 \rgt) = \frac{2x}{2 - x} \\
    f(x, -\sfrac{3}{2}) &= 3 \cdot \lft(1 - \frac{1}{\sqrt{ 1 + \sfrac{2x}{3} }} \rgt)\,, &\quad\quad f(x, \sfrac{3}{2}) &= \frac{3}{2} \cdot \lft( \frac{1}{\lft( 1 - \sfrac{x}{3} \rgt)^{2}} - 1 \rgt) \\
    \powerfun(x, -\sfrac{1}{2}) &= \sqrt{2x+1} - 1\,, &\quad\quad  \powerfun(x, \sfrac{1}{2}) &= \frac{1}{2} \cdot \lft( \lft( x + 1 \rgt)^{2} - 1 \rgt)
\end{align}
Several of these special cases of $f(x, \lambda)$ correspond to curves and transforms that are commonly used in a variety of contexts. The $\lambda = 0$ case is the identity, the $\lambda = -\sfrac{1}{2}$ case is a ``padded'' square root, and the $\lambda=\sfrac{1}{2}$ case is a shifted and scaled quadratic function. The $+\infty$ and $\-1$ cases are both instances of $\operatorname{log1p}$ and the $+1$ and $-\infty$ cases are both instances of $\operatorname{expm1}$.  The $\lambda=-2$ and $\lambda=-3$ cases resemble ``padded'' versions of inverse and inverse-square functions commonly used in physics and geometry.
When $\lambda \in (-\infty, 1)$ the transform is a normalized version of the Box-Cox transform (see Section~\ref{sec:boxcox}). The $\lambda = -2$ case is commonly used in tone mapping~\cite{reinhard2002photographic}.

\paragraph{}
Here is a visualization of this family of transformations, where values of $\lambda$ with simple forms have been annotated.
\vspace{-0.05in}\begin{center}
\hspace{0.43in}\includegraphics[width=0.5\textwidth]{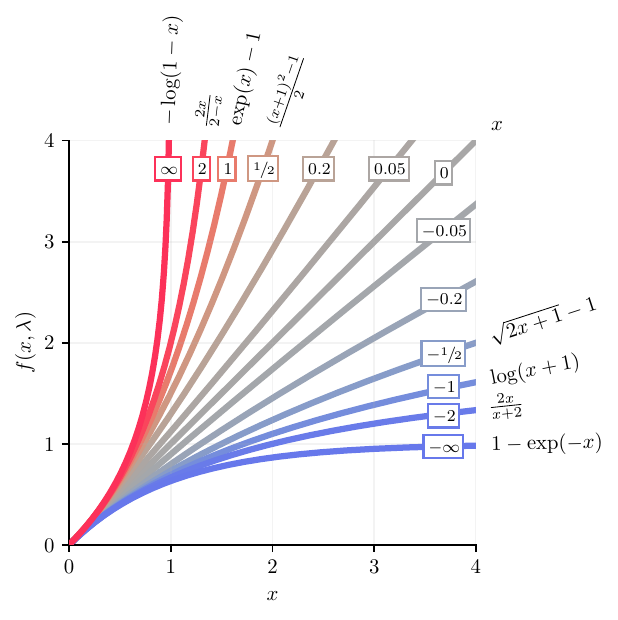}
\end{center}\vspace{-0.15in}

This transform is monotonic with regards to both $x$ and $\lambda$, and is convex when $\lambda > 0$ and concave when $\lambda < 0$.
It is its own inverse: $\powerfun^{-1}(x, \power) = \powerfun(x, -\power)$. At $x=0$ its derivative is 1 and its second derivative is equal to $\operatorname{sign}(\lambda)$, and the same is true for its inverse. Here is the range and domain of $f(x, \lambda)$, if we only consider non-negative inputs:
\begin{equation}
0 \leq x <
\left\{ \begin{array}{cl}
1 & \mathrm{if} \ \power = \infty \\
\frac{\power}{\power-1}  & \mathrm{if} \ 1 < \power < \infty \\
\infty  & \mathrm{if} \ \power \le 1
\end{array} \right.
\,,\quad\quad    
0 \leq \powerfun(x, \power) < \left\{ \begin{array}{cl}
\infty & \mathrm{if} \ \power \ge -1 \\
\frac{\power}{\power + 1} & \mathrm{if} -\infty < \power < -1 \\
1 & \mathrm{if} \ \power = -\infty
\end{array} \right.
\end{equation}
In practice, when calling $f(x, \lambda)$ for values of $\lambda > 1$ it is advisable to clamp $x$ from above to avoid non-finite outputs.

\section{Loss Functions}
\label{sec:loss_function}

Applying $f(x, \lambda)$ to a quadratic (L2) loss function $\frac{1}{2}(\sfrac{x}{c})^2$ with a scale parameter $c > 0$ yields a family of robust losses:
\begin{equation}
\rho\lft(x, \power, c\rgt) \triangleq  \powerfun\lft(\frac{1}{2} \cdot \lft(\sfrac{x}{c}\rgt)^2, \power \rgt)
\end{equation}
Writing out all the removable singularities and limits shows that this family includes L2/quadratic loss itself, and two well-known robust losses: Cauchy (aka Lorentzian) loss~\cite{black1996robust}, and Welsch~\cite{Welsch} (aka Leclerc~\cite{leclerc1989constructing}) loss.
\begin{gather}
\rho\lft(x, \power, c\rgt) = 
\left\{ \begin{array}{@{}l@{\quad}l@{\quad\quad}r} 
-\log \lft( 1 - \frac{1}{2} \cdot \lft( \sfrac{x}{c} \rgt)^{2} \rgt) & \mathrm{if} \ \power = +\infty \\
\exp \lft( \frac{1}{2} \cdot \lft( \sfrac{x}{c} \rgt)^{2} \rgt) - 1 & \mathrm{if} \ \power = 1 \\
\power \cdot \lft( \lft( 1 - \frac{\power - 1}{2 \power} \cdot \lft( \sfrac{x}{c} \rgt)^{2} \rgt)^{\frac{1}{1 - \power}} - 1 \rgt) & \mathrm{if} \ 0 < \power < +\infty \land \power \ne 1 \\
\frac{1}{2} \cdot \lft( \sfrac{x}{c} \rgt)^{2} & \mathrm{if} \ \power = 0 & \text{L2 / Quadratic Loss} \\
\frac{\power}{\power + 1} \cdot \lft( 1 - \lft( 1 - \frac{1}{2 \power} \cdot \lft( \sfrac{x}{c} \rgt)^{2} \rgt)^{\power + 1} \rgt) & \mathrm{if} \ -\infty < \power < 0 \land \power \ne -1  &  \\
\log \lft( 1 + \frac{1}{2} \cdot \lft( \sfrac{x}{c} \rgt)^{2} \rgt) & \mathrm{if} \ \power = -1  & \text{Cauchy / Lorentzian Loss } \\
1 - \exp \lft( - \frac{1}{2} \cdot \lft( \sfrac{x}{c} \rgt)^{2} \rgt) & \mathrm{if} \ \power = -\infty & \text{Welsch / Leclerc Loss}
\end{array} \right.
\end{gather}
Two other well-known robust losses, Charbonnier~\cite{charbonnier1994two} and Geman-McClure Loss~\cite{gemanmcclure}, are also special cases of this family:
\begin{align}
    && \rho(x, -\sfrac{1}{2}, c) &= \sqrt{(\sfrac{x}{c})^2 + 1} - 1 & \text{Charbonnier Loss}  && \\
    && \rho(x, -2, c) &= \frac{2 x^2}{4c^2 + x^2} & \text{Geman-McClure Loss} &&
\end{align}
\vspace{0.05in}
Here is a visualization of this family of robust losses:
\vspace{-0.1in}\begin{center}
\hspace{0.55in}\includegraphics[width=0.9\textwidth]{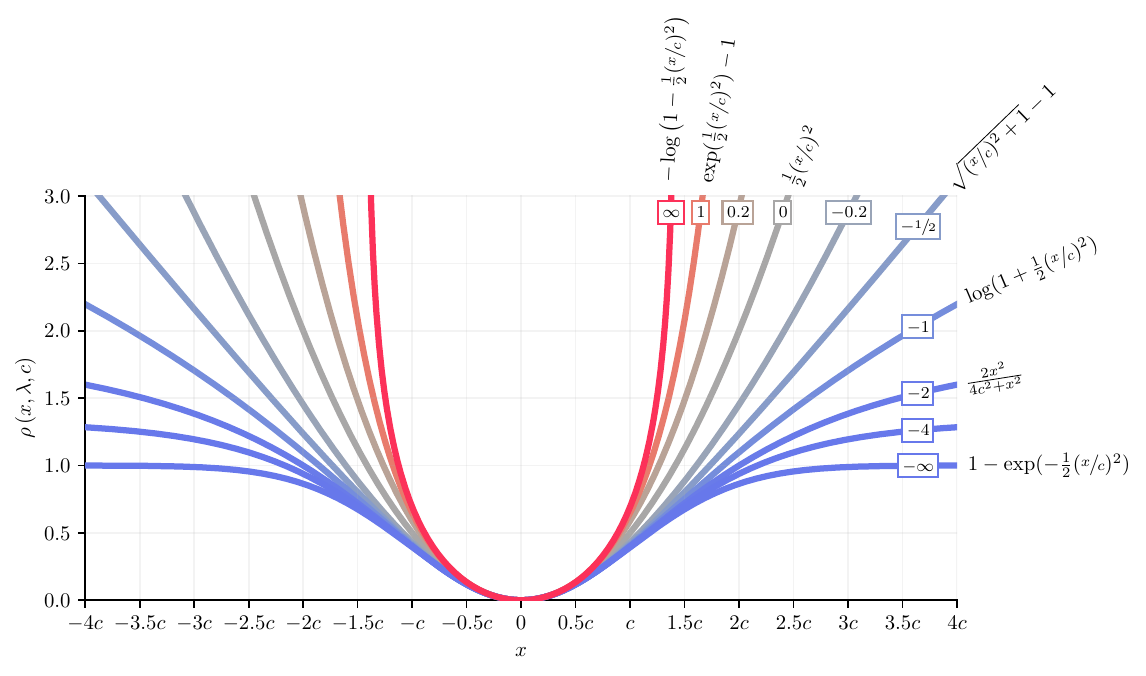}
\end{center}\vspace{-0.15in}
As $\lambda$ decreases from $0$ to $-\infty$, this loss function becomes more robust --- it penalizes outliers less, the gradient of those outliers goes to zero, and estimation of model parameters becomes less sensitive to those outliers. This family of loss functions is a slight superset of a previous family of robust losses I explored~\cite{barron2019general}, which corresponds to this family where $\lambda \leq 1$. See that paper for a discussion of the value of robust losses with variable or adaptive degrees of robustness. This family is also a superset of the ``Smooth Exponential Family'' of loss functions~\cite{tarel2002using}.

\section{Kernel Functions and Iteratively Reweighted Least Squares}
\label{sec:kernel_function}

Taking the derivative of the loss function $\rho(x, \lambda, c)$ and scaling it by $\sfrac{c^2}{x}$ yields a family of stationary kernel functions:
\begin{equation}
    k(x, \power, c) \triangleq \frac{c^2}{x} \cdot \frac{\partial}{\partial x} \rho\lft(x, \power, c \rgt) \label{eq:kernel_unstable}
\end{equation}
Here $x$ is the Euclidean distance between two points.
Equation~\ref{eq:kernel_unstable} is numerically unstable near the origin due to the division by $x$, an so should instead be computed by defining the derivative of $f(x, \lambda, c)$ and evaluating that on $(\sfrac{x}{c})^2$ directly:
\begin{equation}
    k(x, \power, c) = \gradfun \lft( \frac{1}{2} \cdot (\sfrac{x}{c})^2, \power \rgt)\,, \quad\quad g(x, \lambda) = \frac{\partial}{\partial x} f\lft(x, \power, \rgt)
\end{equation}
Writing out the removable singularities and limits of $k(x, \lambda, c)$ shows that several cases are commonly-used kernels~\cite{scholkopf2018learning}.
\begin{equation}
k(x, \power, c) = \left\{ \begin{array}{ll@{\quad\quad}r}
\frac{2 c^{2}}{2 c^{2} - x^{2}} & \mathrm{if} \ \power = +\infty \\
\exp \lft( \frac{1}{2} \cdot \lft( \sfrac{x}{c} \rgt)^{2} \rgt) & \mathrm{if} \ \power = 1 \\
\lft( \frac{\sfrac{1}{\power} - 1}{2} \cdot \lft( \sfrac{x}{c} \rgt)^{2} + 1 \rgt)^{\frac{\power}{1 - \power}} & \mathrm{if} \ 0 < \power < \infty \land \power \ne 1 \\
1 & \mathrm{if} \ \power = 0  \\
\lft( 1 - \frac{1}{2 \power} \cdot \lft( \sfrac{x}{c} \rgt)^{2} \rgt)^{\power} & \mathrm{if} \ -\infty < \power < 0 \land \power \ne -1 & \text{Rational Quadratic Kernel} \\
\frac{2 c^{2}}{2 c^{2} + x^{2}}, & \mathrm{if} \ \power = -1 & \text{Inverse Kernel} \\
\exp \lft( -\frac{1}{2} \cdot \lft( \sfrac{x}{c} \rgt)^{2} \rgt) & \mathrm{if} \ \power = -\infty & \text{Gaussian / RBF Kernel}
\end{array} \right.
\end{equation}
There are more previously-explored kernels within this family that do not need to be special-cased:
\begin{align}
    && k(x, +\sfrac{1}{2}, c) &= 1 + \frac{1}{2} (\sfrac{x}{c})^2  & \text{Quadratic Kernel} && \\
    && k(x, +\sfrac{1}{3}, c) &= \sqrt{1 + (\sfrac{x}{c})^2} & \text{Multiquadric Kernel} && \\
    && k(x, -\sfrac{1}{2}, c) &=  \frac{1}{\sqrt{1 + (\sfrac{x}{c})^2}} & \text{Inverse Multiquadric Kernel} &&
\end{align}
Additionally, when $\lambda < -\sfrac{1}{2}$, this kernel is equivalent to a rescaled Student's t-distribution, so this family of kernel functions can be viewed as a superset of a (non-normalized) Student's t-distribution.

\vspace{0.05in}
Here's a visualization of this family of kernel functions for different values of $\lambda$.
\begin{center}
\hspace{0.35in}\includegraphics[width=0.85\textwidth]{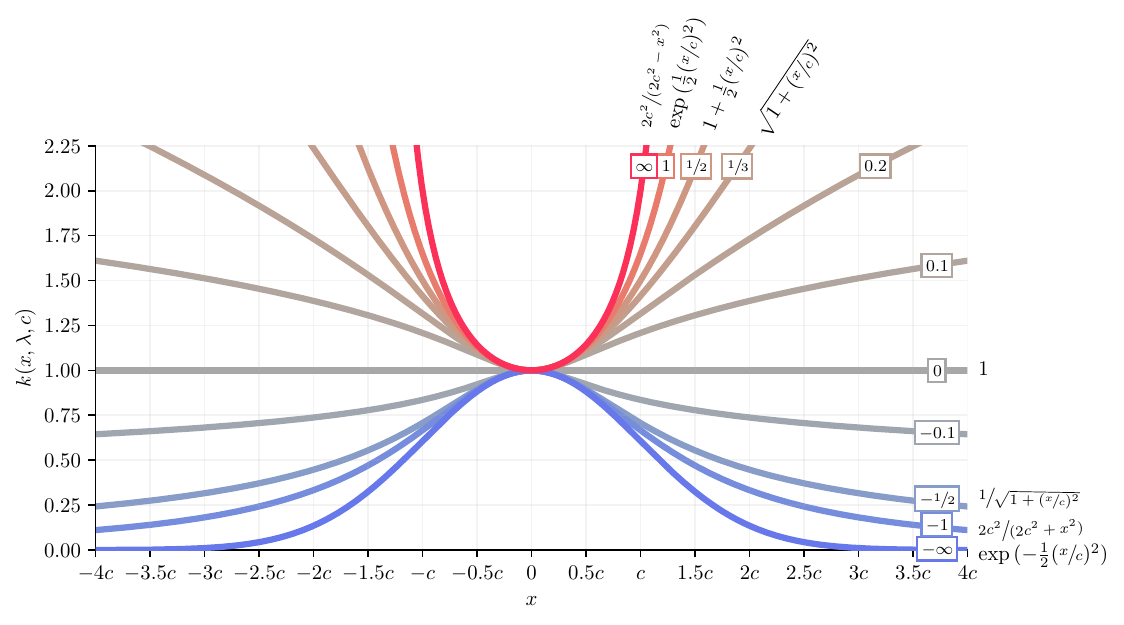}
\end{center}

This kernel, ignoring the $c^2$ scaling, is equivalent to the weights used by iteratively reweighted least-squares~\cite{hampel2011robust} when minimizing $\rho(x, \lambda, c)$.
Anchoring $k(x, \lambda, c)$ to $\rho(x, \lambda, c)$  this way enables some interesting connections. For example: Repeatedly performing least-squares regression with an inverse multiquadric kernel $k(x, -\sfrac{1}{2}, c)$ minimizes Charbonnier loss $\rho(x, -\sfrac{1}{2}, c)$. It is unclear to me if this observation has any practical value, but it is perhaps surprising that these two distant concepts connect in this way.

\section{Probability Distributions}
\label{sec:prob}

% \todo{Call this "A Generalized Gaussian-like Probability Distribution?}

Exponentiating the loss function $\rho(x, \lambda, c)$ and normalizing by the integral yields a family of probability distributions:
\begin{equation}
    \probfun(x, \power, c) = \frac{1}{c \cdot Z(\power)} \exp(-\rho\lft(x, \power, c\rgt))\,, \quad\quad Z(\power) = \int_{-\infty}^{\infty} \exp(-\rho\lft(x, \power, 1\rgt)) \, dx
\end{equation}
% This family includes several PDFs with somewhat straightforward 
This family includes several existing probability distributions:
\begin{align}
\probfun(x, \infty, c) &= \frac{3}{4 c \sqrt{2}} \cdot \operatorname{max}\lft(0, 1 - \frac{1}{2} \lft( \sfrac{x}{c} \rgt)^{2} \rgt) & \text{Epanechnikov Distribution~\cite{Epanechnikov}} \\
% \probfun(x, \sfrac{1}{2}, c) &= \frac{1}{c e^{-\sfrac{1}{4}} K_{\sfrac{1}{4}}(\sfrac{1}{4})} \cdot \exp \lft( -\frac{1}{2} \lft( 1 + \frac{1}{2} \lft( \sfrac{x}{c} \rgt)^{2} \rgt)^{2} \rgt) \\
\probfun(x, 0, c) &= \frac{1}{c \sqrt{ 2 \pi }} \cdot \exp \lft( -\frac{1}{2} \lft( \sfrac{x}{c} \rgt)^{2} \rgt)  & \text{Normal Distribution}\\
\probfun(x, -\sfrac{1}{2}, c) &= \frac{1}{2 c K_1(1)} \cdot \exp \lft(- \sqrt{ 1 + \lft( \sfrac{x}{c} \rgt)^{2} } \rgt) & \text{``Smooth Laplace'' Distribution} \\
\probfun(x, -1, c) &= \frac{1}{c \pi  \sqrt{ 2 } \lft( 1 + \frac{1}{2} \lft( \sfrac{x}{c} \rgt)^{2} \rgt)} & \text{Cauchy Distribution}
\end{align}
The ``smooth Laplace'' distribution approaches the Laplace distribution as $c$ approaches $0$, and is a smooth approximation to it otherwise. $K_v$ is the modified Bessel function of the second kind of real order $v$. The partition function $Z(\lambda)$ does not appear to have a tractable closed form, but can be approximated to within machine epsilon using Simpson's method (see Section~\ref{sec:implementation}). In practice, if $Z(\lambda)$ needs to be evaluated efficiently or differentiably, a precomputed spline or lookup table should be used.

\vspace{0.05in}
Here's a visualization of this family. 
\begin{center}
\includegraphics[width=0.85\textwidth]{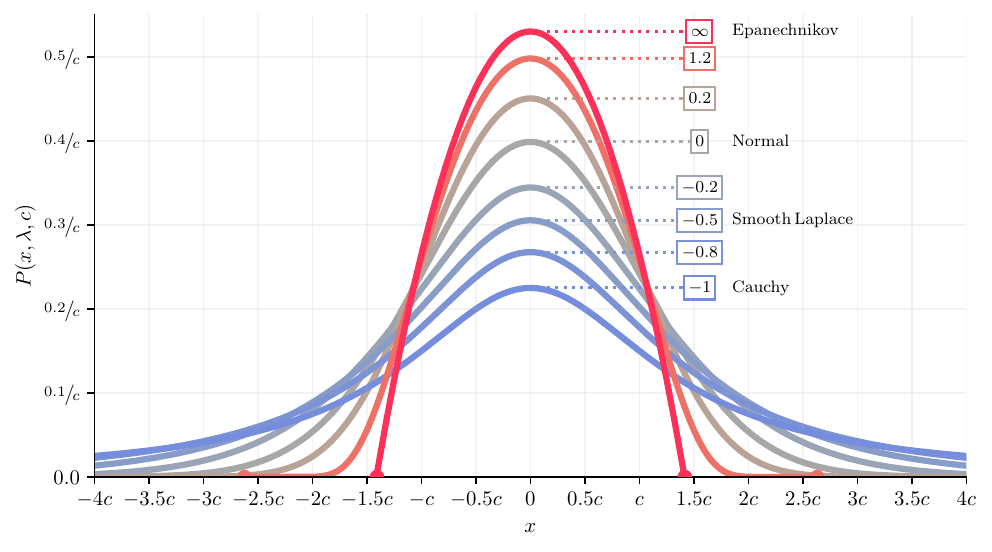}\hspace{0.2in}\phantom{hey}
\end{center}\vspace{-0.15in}
Note that the distribution is undefined when $\lambda < -1$, and that the distribution has a bounded support when $\lambda > 1$ (the extents of which are shown in the plot with markers).

\vspace{0.05in}
This family is similar to Student's t-distribution, in that the family proceeds monotonically from leptokurtic to platykurtic as $\lambda$ increases from $-1$ to $\infty$. That said, the normal and Cauchy distributions appear to be the only points of intersection between the two families. This distribution also resembles Tsallis's \emph{q}-Gaussian~\cite{tsallis1988possible}, though the families appear to be distinct. This distribution is also a superset of the distribution described in my earlier work~\cite{barron2019general} --- they are equivalent when $\lambda \in [-1, 1]$.

% \section{Another Probability Distribution}
% \label{sec:another_prob}

% \todo{Call this "A Student's t-distribution-like Probability Distribution"?}

% \begin{equation}
% Q(x, \power, c) = \frac{f\lft( - \frac{\power}{2} \cdot \lft(\sfrac{x}{c}\rgt)^2, \power\rgt) + \power}{c \cdot Z_Q(\power)}
% \,, \quad\quad 
% Z_Q(\power) = \left\{ \begin{array}{ll}
% \displaystyle \frac{\power \sqrt{ 8 \pi } \cdot \Gamma \lft( \frac{1}{1 - \power} \rgt)}{\lft( 3 - \power \rgt) \sqrt{ 1 - \power } \cdot \Gamma \lft( \frac{3 - \power}{2 - 2 \power} \rgt)} & \mathrm{if} \ \power < 1 \\
% \power \cdot \sqrt{ 2 \pi } & \mathrm{if} \ \power = 1 \\
% \displaystyle \frac{\power \sqrt{ 2 \pi } \cdot \Gamma \lft( \frac{3 - \power}{2 \lft( \power - 1 \rgt)} \rgt)}{\sqrt{ \power - 1 } \cdot \Gamma \lft( \frac{1}{\power - 1} \rgt)} & \mathrm{if} \ \power > 1 \end{array} \right.
% \end{equation}

% \todo{ write, make+add figure. Clarify the range of negative inputs maybe. Call Z in Section 4 Z-sub-p. Update the code block Only defined for p in (0, 3]. This is equivalent to the q-Gaussian, which is a superset of Student's t-distribution}

% % # def Q_z(p):
% % #   p = jnp.asarray(p)
% % #   return switch([
% % #     (p < 1, p * (jnp.sqrt(8 * jnp.pi) * gamma(1 / nozero(1 - p))) / nozero((3 - p) * jnp.sqrt(nozero(1 - p)) * gamma((3 - p) / nozero(2 * (1 - p))))),
% % #     (p == 1, p * jnp.sqrt(2 * jnp.pi)),
% % #     ((p > 1) & (p < 3), p * (jnp.sqrt(2 * jnp.pi) * gamma((3 - p) / nozero(2 * (p - 1)))) / nozero(jnp.sqrt(nozero(p - 1)) * gamma(1 / nozero(p - 1))))
% % #   ], jnp.nan)

\section{Bump Functions}
\label{sec:bump}

% \todo{Try making a bump function using the q-Gaussian approach.}

When $\lambda \in (1, \infty)$, the non-normalized probability distributions shown previously form a family a bump functions~\cite{fry2002smooth}:
\begin{equation}
    b(x, \lambda) = \exp \lft( -f\lft( \frac{\lambda}{\lambda - 1} x^{2}, \lambda \rgt) \rgt)
\end{equation}
Here the input to $f(x, \lambda)$ is rescaled such that all bumps have a support of $[-1, 1]$. This family simplifies to:
\begin{equation}
b(x, \lambda) = \lft\{ \begin{array}{cl}
\exp\lft(\lambda \cdot \lft(1 - \lft(1 - x^2 \rgt)^{\frac{1}{1 - \lambda}} \rgt) \rgt) & \mathrm{if} \, |x| < 1 \\
0 & \mathrm{otherwise}
\end{array} \rgt.
\end{equation}
The ``classic'' bump function is not quite a member of this family, but a scaled and squared version of it is:
\begin{equation}
    b_\text{classic}(x) = \exp \lft( -\lft( \frac{1}{1 - x^{2}} \rgt) \rgt) \cdot \lft( \lft| x \rgt| < 1 \rgt)\,, \quad \quad b(x, 2) =  \lft(e \cdot b_\text{classic}(x) \rgt)^2\,.
\end{equation}
As $\lambda$ approaches $1$ from above, $b(x, \lambda)$ approaches the Dirac delta:
\begin{equation}
\lim_{\lambda \rightarrow 1^+} b(x, \lambda) = 
\begin{cases}
0, & x \neq 0 \\
1, & x = 0
\end{cases}
\end{equation}
And as $\lambda$ approaches $\infty$, the function's smoothness near $x=\pm 1$ decreases as higher order derivatives begin to oscillate, until it converges to an Epanechnikov kernel at $\lambda = \infty$~\cite{Epanechnikov}.

\vspace{0.05in}
Here's a visualization of this family for different valid values of $\lambda$.
\vspace{-0.1in}\begin{center}
\includegraphics[width=0.83\textwidth]{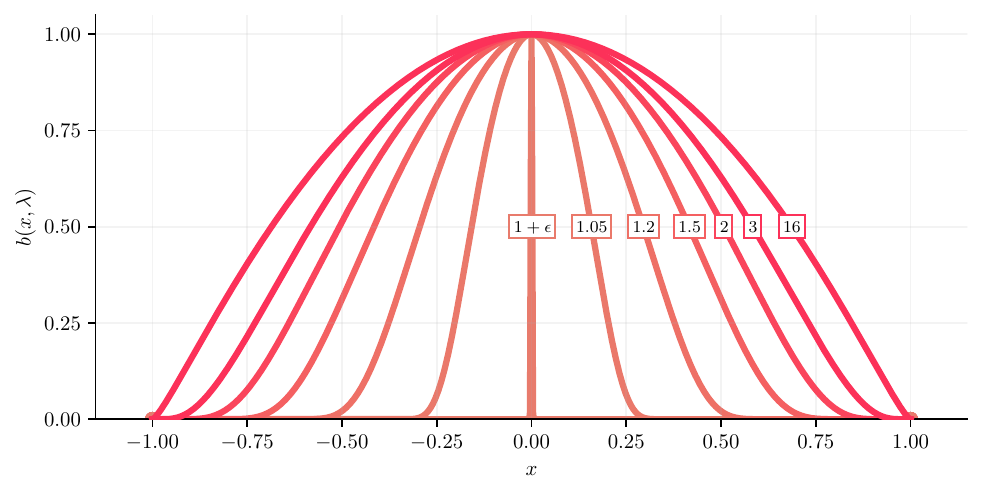}\hspace{0.25in}\phantom{hey}
\end{center}\vspace{-0.15in}

\section{Negative Inputs and Activation Functions}
\label{sec:neg_act}

$f(x, \lambda)$ can be straightforwardly generalized from non-negative inputs to all real numbers, similarly to how the Yeo-Johnson transform~\cite{yeo2000new} generalizes the Box-Cox transform. Because the first and second derivatives of $f(x, \lambda)$ are $1$ and $\pm 1$ at $x=0$ respectively, simply introducing a free parameter $\lambda_-$ for negative inputs (and renaming $\lambda$ to $\lambda_+$) yields an expressive two-parameter family of functions that operate on signed inputs:
\begin{equation}
f_\signed(x, \lambda_+, \lambda_-) = \left\{ \begin{array}{rl}
f \lft( x, \lambda_+ \rgt) & \mathrm{if} \ x \ge 0 \\
-f \lft( -x, \lambda_- \rgt) & \mathrm{if} \ x < 0 \\
\end{array} \right.
\end{equation}
Let's visualize $f_\signed(x, \lambda_+, \lambda_-)$ for different values of $\lambda_+$ and $\lambda_-$:
\begin{center}
\includegraphics[trim={0 .125in 0 .1in}, clip, width=4.85in]{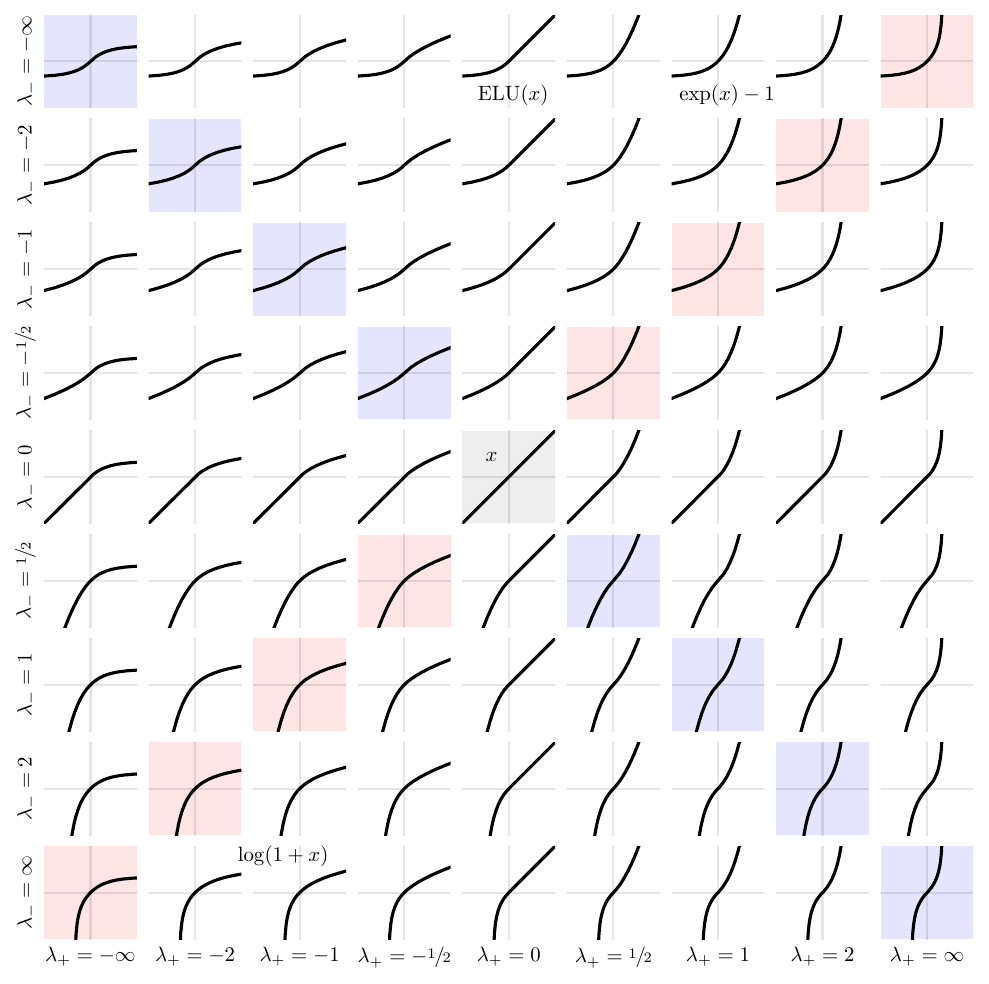}
\end{center}
The upper left quadrant is sigmoid-shaped, the lower right quadrant is logit-shaped, the lower left quadrant is ``log-shaped'', and the upper right quadrant is ``exp-shaped.'' $\exp(x) - 1$, $\log(1 + x)$, and the ELU activation~\cite{clevert2015fast} are members of this family, as shown.
If a single shape parameter is needed, two natural choices are setting $\lambda_- = \lambda_+$, which yields $\operatorname{sgn}(x) \cdot f(|x|, \lambda)$ (shown in blue), and setting $\lambda_- = -\lambda_+$ (shown in red). 
For comparison, here is a visualization of the Yeo-Johnson transform, which is also a single-parameter power transform for signed inputs:
\begin{center}
\includegraphics[trim={0 .125in 0 .1in}, clip, width=4.85in]{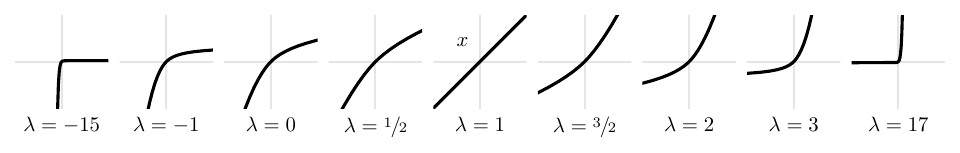}
\end{center}
It resembles $f_\pm(x, \lambda, -\lambda)$ but with non-normalized derivatives at $x=0$, which causes degeneracy when $|\lambda| \gg 1$.

\vspace{0.05in}
As has been noted, many sigmoid-shaped functions can be expressed as the composition of the Box-Cox transform and its inverse~\cite{garcia2005unifying}, and this approach can similarly be used to ground the softplus, sigmoid, and tanh activations in terms of $f(x, \lambda)$:
\begin{equation}
\begin{array}{rll}
\operatorname{softplus}(x) =& f_\signed \lft( f_\signed \lft( x, 1, -\infty \rgt) + 1, -1, \cdot \rgt) &= f \lft(1 +  f \lft( x, 1 \rgt), -1 \rgt)
\\
\operatorname{sigmoid}(x) =& \frac{1}{2} \cdot f_\signed \lft( f_\signed \lft( x + \log(2), 1, -\infty \rgt) + 1, -2, \cdot \rgt) &= \frac{1}{2} \cdot f \lft(1 +  f \lft( x + \log(2), 1 \rgt), -2 \rgt) 
\\
\operatorname{tanh}(x) =& \frac{1}{2} \cdot  f_\signed \lft( f_\signed \lft( 2 x, 1, -\infty \rgt), -2, 2 \rgt) &= \frac{1}{2} \cdot f \lft( f \lft( 2 x, 1 \rgt), -2 \rgt) \\
\operatorname{relu}(x) =& 2-f_\signed(f_\signed(2-x, 2, \lambda_-), -2, -\lambda_-) &=
 2-f(f(2-x, 2), -2)
\end{array}
\end{equation}
The middle column uses $f_\pm(x, \lambda_+, \lambda_-)$ to express each activation, and the right column slightly abuses notation and uses $f(x, \lambda)$ where the input is allowed to be negative. The middle column equality for $\operatorname{relu}(x)$ holds for any value of $\lambda_-$. These equalities are not unique; there appears to be multiple ways to reconstruct some activations using $f(x, \lambda)$. Because these equalities use only $f(x, \lambda)$, multiplication, and addition, it must be the case that for every neural network constructed using only weights, biases, and these activation functions there exists an equivalent neural network that uses only weights, biases, and $f(x, \lambda)$. This suggests that $f(x, \lambda)$ may be a useful tool for exploring the space of neural network activation functions.

\section{Relationship to the Box-Cox Transform}
\label{sec:boxcox}
The power transform presented here is related to the Box-Cox transform~\cite{BoxCox1964jrssB}, shown here in its two-parameter form with the second parameter set to $1$ (note that modifying this parameter is equivalent to scaling the input $x$ and rescaling the output):
\begin{equation}
h(x, \lambda) = \left\{ \begin{array}{ll} \log \lft(1 + x \rgt) & \mathrm{if} \ \lambda = 0 \\ \displaystyle \frac{\lft( x + 1 \rgt)^{\lambda} - 1}{\lambda} & \mathrm{otherwise} \end{array} \right.
\end{equation}
Let's rescale $h(x, \lambda)$ such that, at $x=0$, its first derivative is $1$ and its second derivative is $\operatorname{sign}(\lambda - 1)$:
\begin{equation}
\hat{h}(x, \lambda) = \left\{ \begin{array}{ll}
\log \lft( 1 + x \rgt) & \mathrm{if} \ \lambda = 0 \\ 
\lft( \frac{1}{\lambda} - 1 \rgt) \lft( \lft( 1 + \frac{1}{1 - \lambda} x \rgt)^{\lambda} - 1 \rgt) & \mathrm{if} \ \lambda < 1 \\
x & \mathrm{if} \ \lambda = 1 \\
\frac{\lambda - 1}{\lambda} \cdot \lft( \lft( 1 + \frac{1}{\lambda - 1} x \rgt)^{\lambda} - 1 \rgt) & \mathrm{if} \ \lambda > 1 
\end{array} \right.
\end{equation}
This is the power transform that was used in Zip-NeRF~\cite{barron2023zipnerf}.
$f(x, \lambda)$ is simply this $\hat{h}(x, \lambda)$ where the top half of the transform $\lambda > 1$ is replaced with the inverse of the transform for $\lambda < 1$. Additionally, $\lambda$ is shifted by 1 such that $\lambda = 0$ yields the identity, while in the Box-Cox transform it is $\lambda = 1$ that yields the identity. This is done to emphasize that $\lambda$ \emph{is not} an exponent in $f(x, \lambda)$, while it is in $h(x, \lambda)$; $f(x, 3)$ yields a curve that is very unlike raising $x$ to the power of 3, as setting $\lambda = 3$ would in the Box-Cox transform. This shift also makes it easier to reason about the inverse $f^{-1}(x, \lambda) = f(x, -\lambda)$, and induces an aesthetically pleasant symmetry to the math and to the visualizations.

\vspace{0.05in}
Here are plots of the Box-Cox transform $h(x, \lambda)$ (left), the normalized variant $\hat{h}(x, \lambda)$ (center), and $f(x, \lambda)$ (right).
\vspace{-0.05in}\begin{center}
\includegraphics[width=0.32\textwidth]{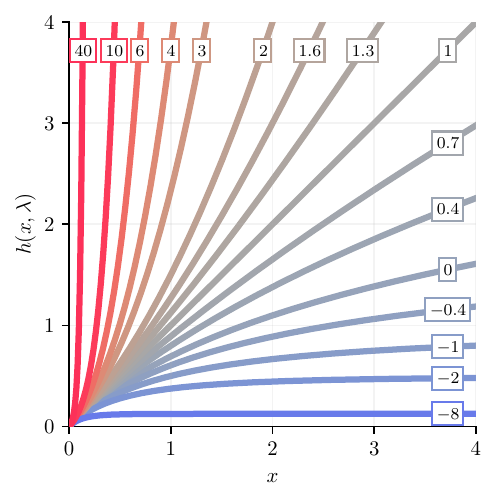}
\,\,
\includegraphics[width=0.32\textwidth]{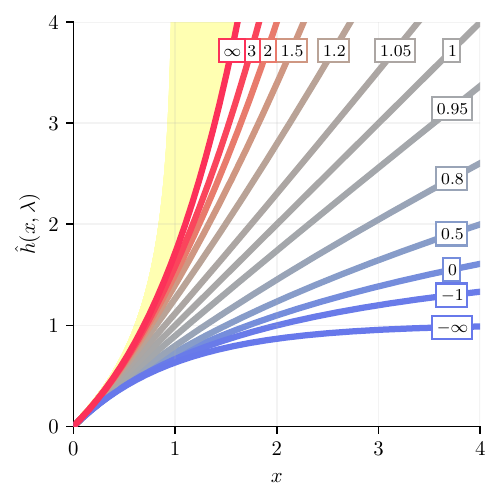}
\,\,
\includegraphics[width=0.32\textwidth]{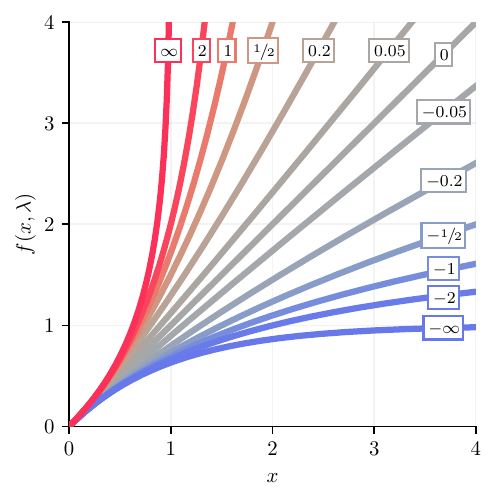}
\end{center}\vspace{-0.15in}
The Box-Cox transform $h(x, \lambda)$ explodes to large (finite) outputs or flattens to near zero as $| \lambda|$ grows, which is likely undesirable behavior for most practitioners.
The yellow region in the plot of $\hat{h}(x, \lambda)$ shows the family of functions that the normalized Box-Cox transform cannot model, which are ``filled in'' by $f(x, \lambda)$.

\vspace{0.05in}
Despite this additional expressive power that $f(x, \lambda)$ has over $h(x, \lambda)$, there exists a bijection between the two transforms:
\begin{equation}
\powerfun(x, \power) = \lft\{ \begin{array}{ll}
 -\lambda \cdot h\lft(-\frac{1}{\lambda} \cdot x, \lambda + 1\rgt) & \mathrm{if} \ \power < 0 \\
 h(x, 1) & \mathrm{if} \ \power = 0 \\
 \frac{\lambda}{1-\lambda} \cdot h\lft(\frac{1-\lambda}{\lambda} \cdot x, \frac{1}{1 - \lambda}\rgt)  & \mathrm{if} \ \power > 0 \end{array} \rgt.,
\quad\quad
h(x, \power) = \lft\{ \begin{array}{ll}
\frac{1}{1 - \power} \cdot f((1-\power) \cdot x, \power-1) & \mathrm{if} \ \power < 0 \\
f(x, 0)  & \mathrm{if} \ \power = 1 \\
 \frac{1}{\power - 1} \cdot f\lft((\power - 1) \cdot x, 1 - \frac{1}{\power}\rgt)  & \mathrm{if} \ \power > 0
\end{array} \rgt.
\end{equation}
This may be somewhat counter-intuitive --- how can a bijection between the two exist if $f(x, \lambda)$ has more expressive power than $\hat{h}(x, \lambda)$? To understand this, note that when $\lambda > 1$, $f(x, \lambda)$ is reducible to $a \cdot ((b \cdot x+1)^q - 1)/q)$, where $q \in [-1, 0)$, and $a$ and $b$ are both negative. The specific negative value of $b$ causes $bx + 1$ to be a \emph{decreasing} linear ramp from 1 to 0 over the domain. This downward sloping ramp then gets raised to a \emph{negative} power $q$ (\ie, it is inverted), which causes $(b \cdot x+1)^q$ to approach infinity as $x$ approaches $1$. The sign of that transformed value is then negated twice, first from the division by $q < 0$, and then from the division by $a < 0$, yielding a positive final value.

\section{Numerical Stability}
\label{sec:numerical}

$f(x, \lambda)$ can and should be implemented using the $\operatorname{expm1}(x) = \exp(x) - 1$ and $\operatorname{log1p}(x) = \log(1 + x)$ functions common in numerical computing packages.
Here is Equation~\ref{eq:f_root} rewritten to use $\operatorname{expm1}(x)$ and $\operatorname{log1p}$:
\begin{equation}
\powerfun(x, \power) = \frac{2 \lft| \power \rgt|\,\,\,}{2 - \lft| \power \rgt| + \power} \cdot \operatorname{expm1} \lft( \lft( 1 - \lft| \power \rgt| \rgt)^{-\operatorname{sign} \lft( \power \rgt)} \cdot \operatorname{log1p} \lft( \frac{2 - \lft| \power \rgt| - \power}{2 \lft| \power \rgt|\,\,\,} x \rgt) \rgt)
\end{equation}
And here are the rewritten cases of Equation~\ref{eq:f_unstable}. Note that $1 - \exp(-x) = -\operatorname{expm1}(-x)$ and $-\log(1-x) = -\operatorname{log1p}(-x)$.
\begin{equation}
\powerfun(x, \power) = \lft\{ \begin{array}{c@{\,\,}c@{}c@{\,\,}c@{\,}c@{\,\,}ll}
-1 & & & \operatorname{log1p}(& -1 & x) & \mathrm{if} \ \power = +\infty \\
& \operatorname{expm1}(& & & & x) & \mathrm{if} \ \power = 1 \\
\phantom{-}\power  & \operatorname{expm1}\!\Big( & \frac{1}{1-\power}   & \operatorname{log1p}\!\big(& \frac{1-\power}{\power} & x \big) \Big) & \mathrm{if} \ 0 < \power < +\infty \land \power \ne 1 \\
& & & & & x & \mathrm{if} \ \power = 0 \\
-\frac{\power}{\power+1} &  \operatorname{expm1}\!\Big( & (\power+1)   & \operatorname{log1p}\!\big(& -\frac{1}{\power} & x \big) \Big) & \mathrm{if} \ -\infty < \power < 0 \land \power \ne -1 \\
& & & \operatorname{log1p}(& & x) & \mathrm{if} \ \power = -1 \\ 
-1 & \operatorname{expm1}(& & &-1 & x) & \mathrm{if} \ \power = -\infty
\end{array} \rgt.
\label{eq:stablecases}
\end{equation}
This math has been formatted such that scale factors and operators are  horizontally aligned across all cases, which shows how $f(x, \lambda)$ can be implemented using just a single vectorized call to both $\operatorname{expm1}$ and $\operatorname{log1p}$ for all values of $\lambda$.

\vspace{0.05in}
The stable implementation yields significantly more accurate results, especially when $\power$ is near $\pm 1$ and when $|\power| \gg 1$:
\vspace{-0.1in}\begin{center}
\includegraphics[width=6.5in]{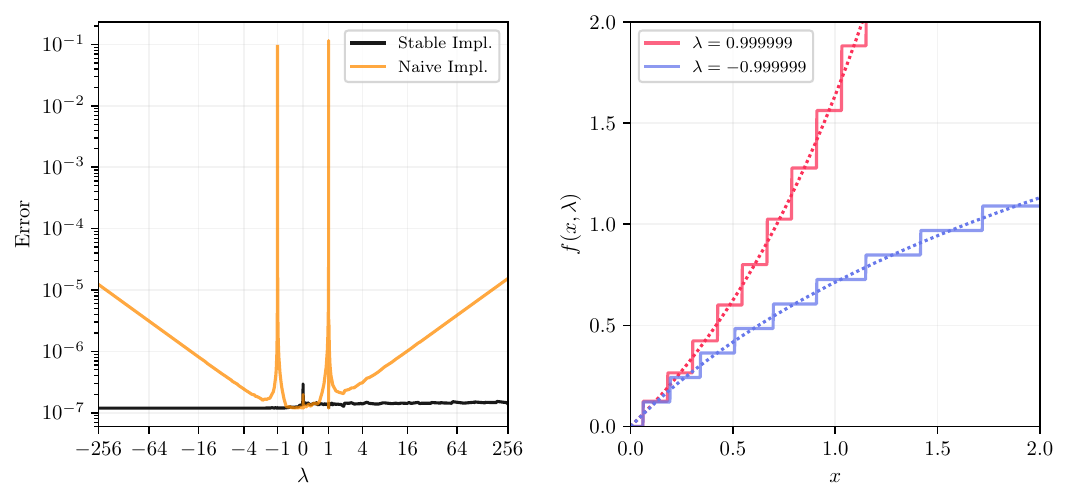}
\end{center}\vspace{-0.15in}
On the left is the error (the geometric mean of the 
absolute error for $x \in [0.01, 1]$) of the naive and stable implementations of $f(x, \power)$ for different values of $\lambda$.
On the right are plots of the stable implementation (dotted lines) and the naive implementation (solid lines) for values of $\power$ close to $\pm 1$. The numerical instability of the naive implementation in these cases takes the form of severe quantization.

\clearpage 

\section{Implementation}

\label{sec:implementation}

Here's a  JAX~\cite{bradbury2018jax} implementation of the stable version of $f(x, \lambda)$ shown in Equation~\ref{eq:stablecases}. The other relevant functions presented in this work are defined in terms of $f(x, \lambda)$. Function and variable names match the notation used in the paper.
\begin{lstlisting}[escapeinside={(*}{*)}]
import jax
import jax.numpy as jnp
import scipy

eps = jnp.finfo(jnp.float32).eps
tiny = jnp.finfo(jnp.float32).tiny

def max_domain((*{\ttfamily $\uplambda$}*)):
  m = jnp.maximum((*{\ttfamily $\uplambda$}*), 1 + eps)/jnp.maximum((*{\ttfamily $\uplambda$}*) - 1, tiny)
  return jnp.nextafter(jnp.where((*{\ttfamily $\uplambda$}*) >= jnp.finfo(jnp.float32).max, 1, m), -jnp.inf)

def f(x, (*{\ttfamily $\uplambda$}*)):
  pinf = (*{\ttfamily $\uplambda$}*) > 1/eps             # Is (*{\ttfamily $\uplambda$}*) practically positive infinity?
  pone = jnp.abs((*{\ttfamily $\uplambda$}*) - 1) < eps  # Is (*{\ttfamily $\uplambda$}*) near positive 1?
  zero = jnp.abs((*{\ttfamily $\uplambda$}*)) < tiny     # Is (*{\ttfamily $\uplambda$}*) near zero?
  none = jnp.abs((*{\ttfamily $\uplambda$}*) + 1) < eps  # Is (*{\ttfamily $\uplambda$}*) near negative one?
  ninf = (*{\ttfamily $\uplambda$}*) < -1/eps            # Is (*{\ttfamily $\uplambda$}*) practically negative infinity?
  ow_pos = ((*{\ttfamily $\uplambda$}*) > 0) & jnp.logical_not(pinf | pone | zero)  # Otherwise, is (*{\ttfamily $\uplambda$}*) positive?
  ow_neg = ((*{\ttfamily $\uplambda$}*) < 0) & jnp.logical_not(ninf | none | zero)  # Otherwise, is (*{\ttfamily $\uplambda$}*) negative?
  nozero = lambda x: jnp.where(jnp.abs(x) < tiny, tiny, x)
  switch = lambda c, d: jnp.select((***)zip((***)c), d)
  # Clip x to f()'s domain to prevent NaNs.
  x = jnp.minimum(x, max_domain((*{\ttfamily $\uplambda$}*)))
  x (***)= switch([(ninf | pinf, -1), (ow_neg, -1/nozero((*{\ttfamily $\uplambda$}*))), (ow_pos, (1 - (*{\ttfamily $\uplambda$}*))/nozero((*{\ttfamily $\uplambda$}*)))], 1)
  x = jnp.where(ninf | zero | pone, x, jnp.log1p(x))
  x (***)= switch([(ow_neg, (*{\ttfamily $\uplambda$}*) + 1), (ow_pos, 1/nozero(1 - (*{\ttfamily $\uplambda$}*)))], 1)
  x = jnp.where(none | zero | pinf, x, jnp.expm1(x))
  x (***)= switch([(ninf | pinf, -1), (ow_neg, -(*{\ttfamily $\uplambda$}*)/nozero((*{\ttfamily $\uplambda$}*) + 1)), (ow_pos, (*{\ttfamily $\uplambda$}*))], 1)
  return x

f_inv = lambda x, (*{\ttfamily $\uplambda$}*): f(x, -(*{\ttfamily $\uplambda$}*))
rho = lambda x, (*{\ttfamily $\uplambda$}*), c = 1: f(0.5 (***) (x/c)(***)(***)2, (*{\ttfamily $\uplambda$}*))
g = lambda x, (*{\ttfamily $\uplambda$}*): jnp.vectorize(jax.grad(lambda z: f(z, (*{\ttfamily $\uplambda$}*))))(jnp.minimum(x, max_domain((*{\ttfamily $\uplambda$}*))))
k = lambda x, (*{\ttfamily $\uplambda$}*), c = 1: g(0.5 (***) (x/c)(***)(***)2, (*{\ttfamily $\uplambda$}*))
P = lambda x, (*{\ttfamily $\uplambda$}*), c = 1: jnp.exp(-rho(x, (*{\ttfamily $\uplambda$}*), c)) / (c (***) Z((*{\ttfamily $\uplambda$}*)))

def Z((*{\ttfamily $\uplambda$}*), num_simpson=2048):
  assert (*{\ttfamily $\uplambda$}*) >= -1
  # Find the x where non-normalized P(x, (*{\ttfamily $\uplambda$}*), 1) drops below epsilon.
  x_max = jnp.sqrt(2 (***) f_inv(-jnp.log(eps(***)(***)2), (*{\ttfamily $\uplambda$}*)))
  # Sample points from [0, x_max] curved by f(x, 1) and its inverse.
  x = f(jnp.linspace(0, f_inv(x_max, 1), num_simpson), 1)
  # Get un-normalized probabilities.
  y = jnp.exp(-f(1/2 (***) x(***)(***)2, (*{\ttfamily $\uplambda$}*)))
  # Approximate the integral, double it because we only sampled half the support.
  z = 2 (***) scipy.integrate.simpson(y, x=x)
  return z

def b(x, (*{\ttfamily $\uplambda$}*)):
  assert (*{\ttfamily $\uplambda$}*) > 1 and (*{\ttfamily $\uplambda$}*) < jnp.inf
  return jnp.exp(-f((1 / (1 - 1/(*{\ttfamily $\uplambda$}*))) (***) x(***)(***)2, (*{\ttfamily $\uplambda$}*)))

f_pm = lambda x, (*{\ttfamily $\uplambda$}*)_pos, (*{\ttfamily $\uplambda$}*)_neg: jnp.where(x >= 0, f(x, (*{\ttfamily $\uplambda$}*)_pos), -f(-x, (*{\ttfamily $\uplambda$}*)_neg))
\end{lstlisting}

% (***)
% (*{\ttfamily $\uplambda$}*)

\clearpage

{
\small
\bibliographystyle{ieeenat_fullname}
\bibliography{main}
}

\end{document}

%% file: preamble.tex
\usepackage{xfrac}

\usepackage[dvipsnames]{xcolor}

\usepackage{cmtt}
\usepackage{upgreek}
\usepackage{fix-cm}  % Allows arbitrary font sizes

\newcommand\lft{\mathopen{}\left}
\newcommand\rgt{\aftergroup\mathclose\aftergroup{\aftergroup}\right}

\newcommand{\powerfun}{f}
\newcommand{\gradfun}{g}
\newcommand{\power}{\lambda}

\newcommand{\probfun}{P}

\newcommand{\signed}{\pm} %{\mathrm{signed}}